\documentclass[acmtog]{acmart}

\AtBeginDocument{%
  
}

\usepackage{float}
\usepackage{enumitem}
\usepackage{bbm}

\graphicspath{{figures/}}

\copyrightyear{2026}
\acmYear{2026}
\setcopyright{cc}
\setcctype{by}
\acmConference[SA Conference Papers '26]{SIGGRAPH Asia 2026 Conference Papers}{December 01--04, 2026}{Kuala Lumpur, Malaysia}
\acmBooktitle{SIGGRAPH Asia 2026 Conference Papers (SA Conference Papers '26), December 01--04, 2026, Kuala Lumpur, Malaysia}
\acmDOI{10.1145/3829340.3842340}
\acmISBN{979-8-4007-2842-6/2026/12}

\title{Proximity3D: Shape from Capacitive Proximity on Sensing Manifold}

\newcommand{\cuhkcpiiaffiliation}{%
  \affiliation{%
    \institution{Centre for Perceptual and Interactive Intelligence}
    \country{Hong Kong, China}
  }%
}

\author{Hao Chen}
\authornote{Joint first authors.}
\affiliation{%
  \institution{Centre for Perceptual and Interactive Intelligence / Xiamen University}
  \country{China}
}

\author{Chenming Wu}
\authornotemark[1]
\affiliation{%
  \institution{Independent Researcher}
  \country{China}
}

\author{Chun Ping Lam}
\cuhkcpiiaffiliation

\author{Xiangjia Chen}
\cuhkcpiiaffiliation

\author{Guoxin Fang}
\affiliation{%
  \institution{The Chinese University of Hong Kong / Centre for Perceptual and Interactive Intelligence}
  \country{China}
}

\author{Charlie C. L. Wang}
\affiliation{%
  \institution{The University of Manchester}
  \city{Manchester}
  \country{United Kingdom}
}

\author{Yeung Yam}
\affiliation{%
  \institution{The Chinese University of Hong Kong / Centre for Perceptual and Interactive Intelligence}
  \country{China}
}

\author{Juncong Lin}
\authornotemark[2]
\affiliation{%
  \institution{Xiamen University}
  \city{Xiamen}
  \country{China}
}
\email{jclin@xmu.edu.cn}

\author{Chengkai Dai}
\authornote{Corresponding authors: ckdai@cpii.hk (Chengkai Dai) and jclin@xmu.edu.cn (Juncong Lin).}
\cuhkcpiiaffiliation
\email{ckdai@cpii.hk}

\renewcommand{\shortauthors}{Chen et al.}
\authorsaddresses{Authors' Contact Information: Hao Chen, \nolinkurl{30920241154545@stu.xmu.edu.cn}, Centre for Perceptual and Interactive Intelligence, Hong Kong SAR of China and Xiamen University, Xiamen, China; Chenming Wu, \nolinkurl{wcm94@live.com}, Independent Researcher, China; Chun Ping Lam, \nolinkurl{cplam@link.cuhk.edu.hk}, Centre for Perceptual and Interactive Intelligence, Hong Kong SAR of China; Xiangjia Chen, \nolinkurl{xjchen@cpii.hk}, Centre for Perceptual and Interactive Intelligence, Hong Kong SAR of China; Guoxin Fang, \nolinkurl{guoxinfang@cuhk.edu.hk}, The Chinese University of Hong Kong / Centre for Perceptual and Interactive Intelligence, Hong Kong SAR of China; Charlie C.L. Wang, \nolinkurl{charlie.wang@manchester.ac.uk}, The University of Manchester, Manchester, United Kingdom; Yeung Yam, \nolinkurl{yyam@mae.cuhk.edu.hk}, The Chinese University of Hong Kong / Centre for Perceptual and Interactive Intelligence, Hong Kong SAR of China; Juncong Lin, \nolinkurl{jclin@xmu.edu.cn}, Xiamen University, Xiamen, China; Chengkai Dai, \nolinkurl{ckdai@cpii.hk}, Centre for Perceptual and Interactive Intelligence, Hong Kong SAR of China.}

\begin{document}

\begin{abstract}
    Most shape reconstruction methods assume measurements defined over planar sensing domains, such as RGB images or depth maps. In this paper, we use a curved capacitive textile as a shape sensor, treating its surface as a non-planar sensing manifold. Each scan is represented as a capacitive proximity field on this manifold, induced by the interaction between the curved electrode layout and nearby object geometry. We introduce a multi-view feedforward reconstruction model that aggregates these fields across known sensor views and recovers the observed object shape. Simulated and physical experiments demonstrate robust reconstruction from capacitive proximity signals acquired on curved sensing surfaces, pointing toward a new route to robotic near-field geometric awareness via embodied sensing.

\end{abstract}



\begin{teaserfigure}
    \centering
    \includegraphics[width=0.9\textwidth]{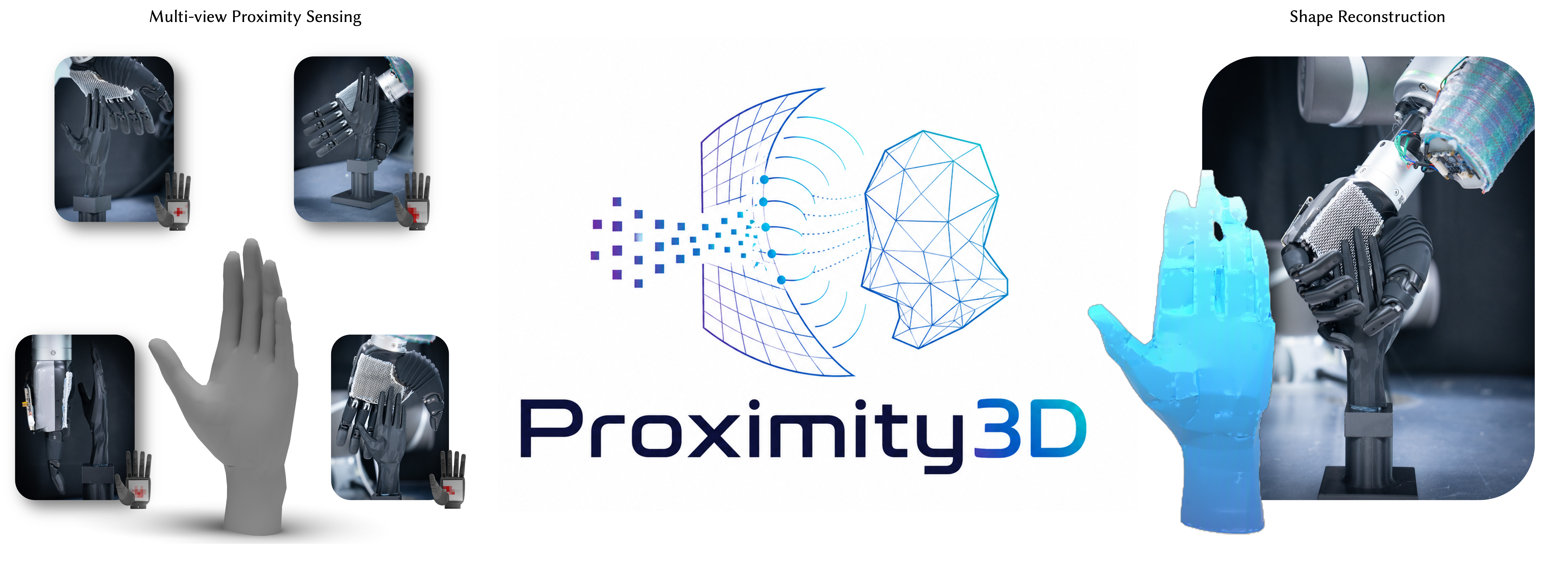}
    \caption{Proximity3D reconstructs object geometry from pre-contact
    capacitive scans captured by a curved woven sensing manifold. As the
    dexterous hand moves the palm-mounted sensing manifold
    (shown as the white region in the left) around the target, each view produces a
    proximity field on the woven surface; our model interprets these posed
    fields on the sensing manifold, fuses their geometric evidence, and
    recovers a 3D mesh, achieving the fidelity necessary to support downstream robotic applications.}
    \Description{A left-to-right overview combining photographs, schematics, and reconstructed geometry. On the left, four photographs show a palm-mounted woven sensor approaching a hand-shaped target from different poses, arranged around the target model. The center schematic links sampled capacitive responses on a curved sensing surface to a polygonal object mesh. On the right, a photograph of the sensor scanning a physical hand is paired with the recovered blue hand geometry, connecting multi-view acquisition to the final reconstruction.}
    \label{fig:teaser}
\end{teaserfigure}

\maketitle

\section{Introduction}

In robotic manipulation, the near-field region—--the narrow gap between a robot and an object prior to contact—---is frequently occluded from line-of-sight observation. This limitation motivates a shift toward \emph{embodied near-field sensing}, where the robot body itself becomes the sensing domain. Non-contact capacitive proximity sensing is well suited to this setting, as nearby conductive objects perturb the electric field around the sensor, thereby modulating the capacitance and generating measurable proximity signals. However, existing approaches typically utilize proximity signals as collision heuristics~\cite{deanleon2017tomm} or simple distance estimates~\cite{mayton2010electric}. Such approaches do not capture richer spatial information encoded in these proximity signals. Also, for applications such as grasp planning and shape-aware manipulation, the robot would benefit from recovering the surface geometry of the target object before contact. Our goal is therefore to recover 3D shape from multi-view capacitive proximity fields captured on a woven sensing manifold conforming to the curved surface, as illustrated in Fig.\ref{fig:teaser}.

The primary challenge lies in the indirect and sparse nature of these capacitive proximity fields. First, the sensor responses are indirect: each channel value reflects complex capacitive coupling between the local electrode layout and nearby object geometry, rather than providing a direct sample of depth or occupancy~\cite{gjoka2025capacitive}. Interpreting these signals, therefore, requires the sensor's own local geometry to be explicitly integrated into the representation. Second, compared to high-resolution image or depth observations, a capacitive scan has significantly lower information density, comprising only a few hundred channel responses due to trace routing density limits in fabrication. These readouts impose a sparse set of indirect constraints on the observed object, making it challenging to recover a complete shape whose geometry remains consistent with such limited data. 

To address these challenges, we propose Proximity3D, a feedforward reconstruction pipeline. Operating directly on the sensing manifold, the network first leverages the local electrode frames and non-uniform electrode layout to perform local aggregation over neighboring channel responses through our proposed \emph{Manifold Sensing Attention} (MSA) operator. The aggregated per-view sensing features are then fused into shape tokens, which are transformed by a global module into a 3D latent representation. With the help of pretrained shape decoder~\cite{xiang2025trellis2} with strong shape prior, our pipeline enables the recovery of complete 3D geometry from sparse capacitive responses, achieving the fidelity necessary to support downstream robotic applications as discussed in Sec.\ref{sec:application}.

Finally, to scale training data beyond physical collection, a surrogate forward model for capacitive proximity fields generation is proposed to adapt local geometry-derived priors into real sensor response, synthesizing realistic proximity fields for unseen meshes and different sensor poses.

Our contributions are as follows:

\begin{itemize}[topsep=0.25em]
    \item We introduce \emph{Proximity3D}, a framework for shape reconstruction
    from multi-view capacitive proximity fields captured on a curved sensing
    manifold. To the best of our knowledge, it is the first method that
    effectively addresses this challenge.
    
    \item We propose \emph{Manifold Sensing Attention} (MSA), an attention mechanism that aggregate each sensing channel's neighbors via their local tangent frames, thereby capturing the underlying relationship of the local sensing context of the sensing manifold.

    \item We propose a surrogate capacitive model that learns realistic capacitive responses from local geometry-derived priors and real sensor response for scalable capacitive proximity fields generation. 

\end{itemize}

\section{Related Work}

\subsection{Sparse and Indirect 3D Reconstruction}
\label{sec:related_sparse}

While traditional 3D reconstruction relies on dense visual or depth data via multi-view stereo~\cite{hartley2003multiple,schoenberger2016sfm,seitz2006comparison,yao2018mvsnet} or volumetric fusion~\cite{curless1996volumetric,newcombe2011kinectfusion,oleynikova2017voxblox}, a variety of learning-based representations such as DeepSDF, Occupancy, Neural Radiance Field (NeRF), and 3D Gaussian Splatting~\cite{park2019deepsdf,mescheder2019occupancy, mildenhall2020nerf, kerbl2023gs} have significantly improved the quality and precision of 3D reconstruction even with sparse inputs. Recent neural reconstruction models further extend these concepts to robust feedforward single- or multi-view inference~\cite{hong2023lrm,wang2024pflrm}. However, these methods presuppose standard inputs of multi-view images from cameras. In contrast, our observations consist of capacitive proximity fields defined directly on a physical manifold surface.

Tactile and visuo-tactile reconstruction deals with similarly localized and sparse data. Elastomer-based optical tactile sensors (e.g., GelSight~\cite{yuan2017gelsight}, DIGIT~\cite{lambeta2020digit}) capture local surface deformations, which can be integrated into global shapes using implicit fields, touch-conditioned diffusion priors, or neural tracking frameworks~\cite{smith2020shape,smith2021active,comi2023touchsdf,wang2025touch2shape,suresh2024neuralfeels}. Yet, these techniques necessitate physical contact and rely entirely on `depth-like' contacting images. Capacitive proximity sensing, conversely, is non-contact and governed by electrical fringing fields rather than optical projection~\cite{mayton2010electric}. Consequently, the sensor's physical configuration, such as channel positions, local tangent frames, etc., must be explicitly integrated to correctly interpret the signals.

\subsection{Capacitive and Proximity Sensing}
\label{sec:related_skins}

Large-area electronic skins distribute contact, force, and deformation sensing across robot bodies~\cite{dahiya2010tactile,yousef2011tactile,hammock2013evolution,mandil2023tactile}, often utilizing conformable substrates to wrap curved surfaces~\cite{someya2005conformable,mannsfeld2010microstructured,boutry2018bioinspired}. Extending beyond physical contact, capacitive and multimodal arrays enable pre-contact proximity awareness. Applications range from whole-body collision avoidance in collaborative arms~\cite{tsuji2018whole} and bio-inspired spatial tracking~\cite{zhou2024mormyroidea} to flexible platforms like CySkin~\cite{giovinazzo2024cyskin} and wireless multimodal e-skins~\cite{markvicka2020wireless,kwon2022decoupled}. Nevertheless, these systems primarily utilize proximity signals for threshold-based alarms, gesture recognition, or coarse spatial localization. They generally do not attempt the highly ill-posed task of inverting sparse capacitance readings into 3D geometry.

Recent advances in digital fabrication and simulation provide the foundation for such inversion. Computational textiles, driven by geodesic stitch-maps~\cite{liu2021knitting4d} and 3D freeform weaving~\cite{chen2024freeformweaving}, allow sensors to be integrated into prescribed, calibrated surface parameterizations~\cite{DaiFreeformSkin}. Concurrently, forward models for capacitive stretch sensors have been formulated as computable functionals of electrode geometry and material state~\cite{gjoka2025capacitive}, highlighting the tight coupling between sensor readouts and their underlying geometry. While these works establish the physical substrates and forward models, our method tackles the corresponding inverse problem: reconstructing 3D geometry from posed capacitive proximity fields induced across these fabricated manifolds.

\subsection{Geometric Learning and Multi-View Latent Decoders}
\label{sec:related_geometric}

\paragraph{Learning on surfaces.}
Geometric learning adapts neural operators to non-Euclidean domains~\cite{kipf2017semi,velickovic2018graph} and specialized architectures for meshes via edge convolutions, subdivision, or diffusion~\cite{bronstein2017geometric,bronstein2021geometric,hanocka2019meshcnn,hu2022subdivnet,monti2017geometric,sharp2022diffusionnet}. Within robotics, graph-based methods process tactile data from irregular taxel arrays~\cite{garcia2019tactilegcn,gu2020tactilesgnet,jiang2025robot}. We adapt these principles to curved capacitive skins. Specifically, our aggregation stage via manifold sensing attention injects the relative continuous 6D pose~\cite{zhou2019continuity} between neighboring electrode frames directly into the attention mechanism~\cite{vaswani2017attention}. 

\paragraph{Multi-view fusion and structured shape latents.}
Feedforward reconstruction often employs transformers to encode posed observations into global representations like tri-planes or volumetric fields~\cite{hong2023lrm,wang2024pflrm}. Latent-query architectures handle variable-sized inputs by cross-attending from fixed latent queries to variable-length token sets~\cite{jaegle2022perceiver,zhang2023shape2vecset}. More recently, generative priors like TRELLIS~\cite{xiang2024structured} and its successor~\cite{xiang2025trellis2} have introduced structured latent spaces, bridging sparse spatial cells with powerful pretrained mesh decoders. Our shape token reconstruction leverages this structured latent formulation as a fixed output interface. By augmenting posed scan tokens with Fourier pose encodings~\cite{mildenhall2020nerf} and register tokens~\cite{darcet2024registers,oquab2024dinov2}, we fuse multi-view proximity data into a compact scene representation. Learned cell queries then extract the necessary support and per-cell features, allowing our pipeline to exploit a pre-trained generative shape decoder.

\section{Method}
\label{sec:method}

\emph{Proximity3D} maps multi-view capacitive proximity fields captured on a woven sensing manifold to the 3D geometry of a nearby object through a feedforward reconstruction pipeline (Fig.~\ref{fig:method_overview}). During acquisition, the sensing manifold is moved around a conductive object; the object perturbs the fringing electric field around the electrodes, producing one capacitance response at each channel site for every sensor view. In this setting, the readouts are sparse in the number of electrodes and indirect---each channel reports a capacitance value rather than explicit geometric information. The reconstruction network therefore aggregates the readouts at two levels. Local aggregation encodes neighboring channel responses using the electrode layout and electrode tangent frames, while global aggregation integrates per-view features together with their known sensor poses into a 3D latent representation. A shape decoder then reconstructs the final object geometry under the pre-trained shape prior.

The sensing principle and fabrication process are introduced in the supplemental material.

\begin{figure}[t]
    \centering
    \includegraphics[width=0.45\textwidth]{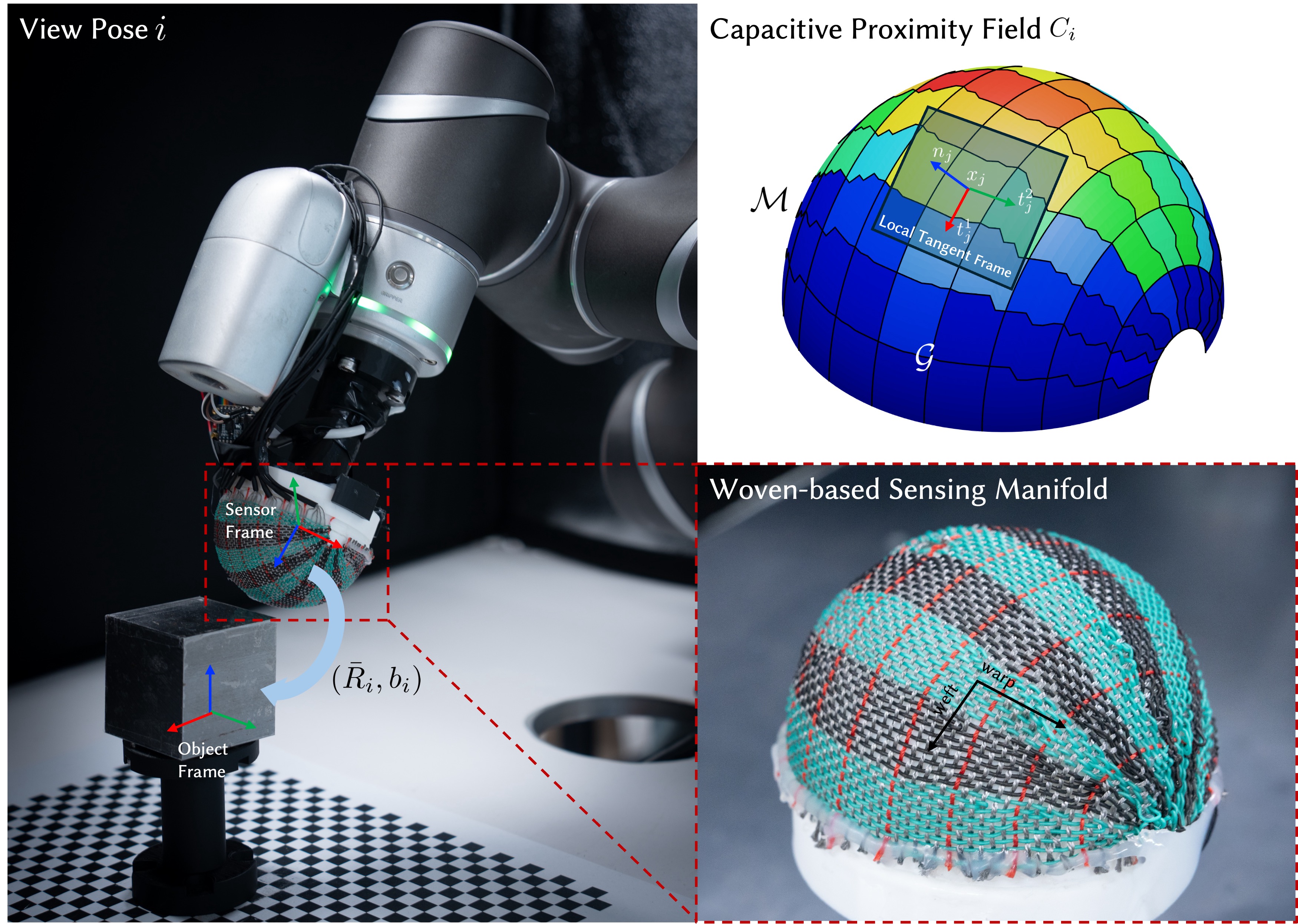}
    \caption{Illustration of the woven-based hemispherical sensing manifold. Highlighting  the warp-weft textile structure that embeds the electrode layout on the sensing surface.}
    \Description{A robotic arm holds a hemispherical woven capacitive sensor
    near a target object. Insets show the sensor frame, object frame, view pose,
    a colored capacitive proximity field on the curved manifold, local tangent
    directions at a channel site, and a close-up of the woven warp-weft
    structure.}
    \label{fig:hemisphere}
\end{figure}

\subsection{Problem Definition}
\label{sec:problem_definition}

We represent the woven sensor as a sensing manifold \(\mathcal M\) as shown in Fig.\ref{fig:hemisphere}, with \(N\) electrode channel sites \(\{x_j\}_{j=1}^{N}\) sampled on \(\mathcal M\), each associated with an orthonormal local electrode tangent frame \(R_j=[t_j^1,t_j^2,n_j]\in SO(3)\) defined by the warp-weft parameterization \cite{chen2024freeformweaving}. The woven layout defines the channel connectivity \(\mathcal G=(\{1,\ldots,N\},E)\), where \(E\) connects neighboring channels in the textile layout. The fabrication and electrical characterization of the woven sensor are not the focus of this work. We therefore treat the sensor's geometric information and connectivity as known inputs to the reconstruction problem.

For each sensor view \(i\), the sensor records a capacitive proximity field
\(C_i=[C_{i1},\ldots,C_{iN}]\in\mathbb R^{N}\), where \(C_{ij}\) is the
normalized response at channel site \(j\) on the sensing manifold. The
associated view pose \((\bar R_i,b_i)\), comprising rotation \(\bar R_i\in SO(3)\) and translation \(b_i\in\mathbb R^3\), maps the sensor frame to the canonical object frame and is derived from hardware kinematics. Given \(V\) feasible views  around the object, the reconstruction pipeline takes \(\{(C_i,\bar R_i,b_i)\}_{i=1}^{V}\) along with the pre-defined sensor geometry
\((\{x_j,R_j\}_{j=1}^{N},\mathcal G)\) defined in the sensor frame, reconstructing an object mesh \(\hat M\) in the canonical object frame.

\subsection{Aggregation via Manifold Sensing Attention}
\label{sec:msa}

The capacitive proximity fields are first processed directly on the sensing manifold $\mathcal M$. An approaching object induces a capacitance proximity field over the sensing manifold. Due to complex capacitive coupling with local neighbors, these channel responses are strongly direction-dependent: the same object shape yields different responses on channel site $j$ depending on its orientation relative to the local electrode frame $R_j$ despite the same relative position. In addition, the fabrication process of the curved woven surface introduces non-uniform electrode spacing and orientation over the sensing surface.

As a result, a proximity response cannot be interpreted independently of its local sensing geometry and connectivity. We therefore aggregate electrode readouts via Manifold Sensing Attention (MSA) in the local electrode tangent frame, so the resulting per-channel features jointly encode the proximity response and the directional dependencies produced by the local electrode layout on the sensing manifold.
\begin{figure*}[t]
    \centering
    \includegraphics[width=\textwidth]{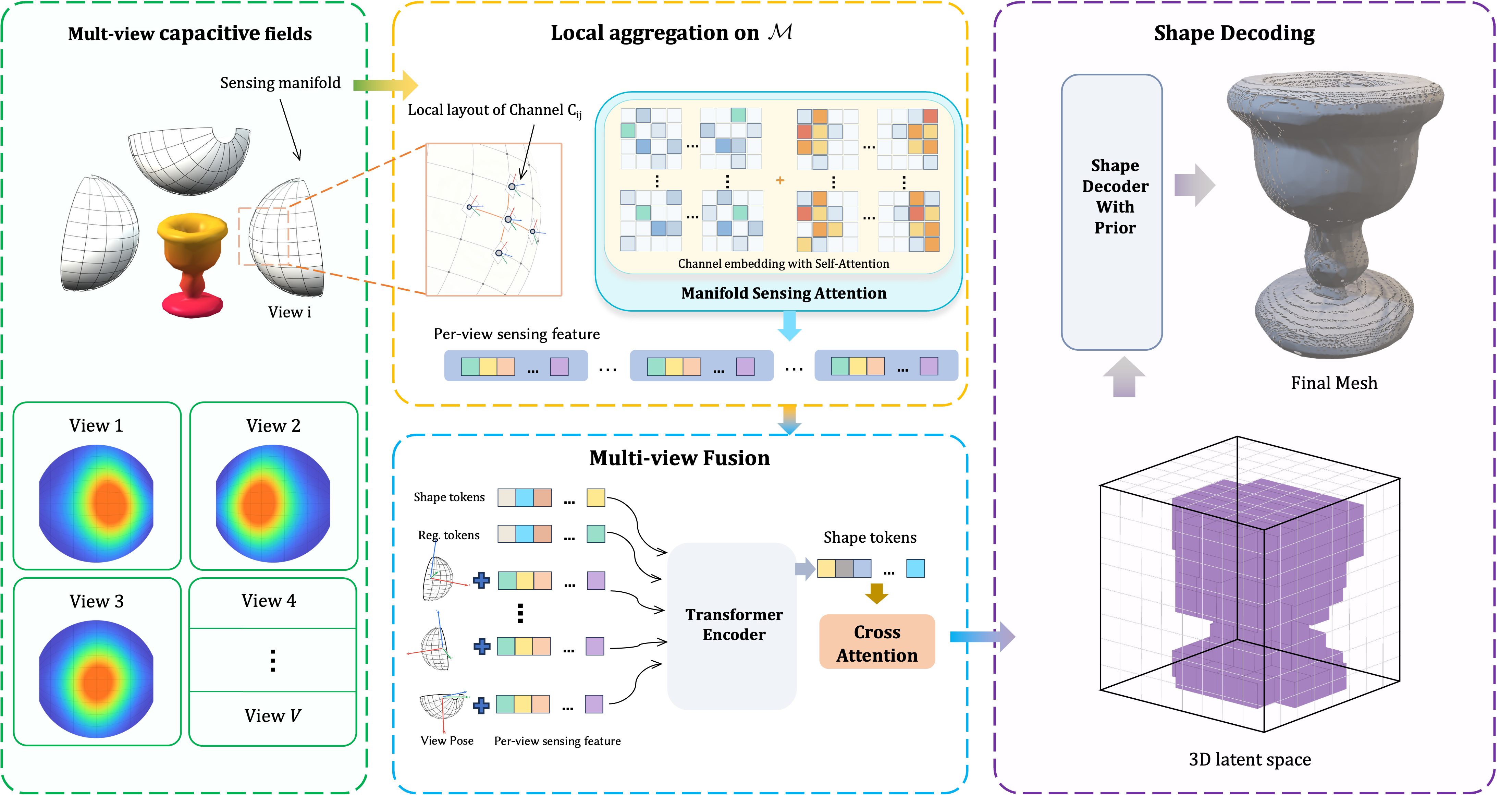}
    \caption{Overview of Proximity3D. Multi-view capacitive fields are sampled
    on a sensing manifold from different sensor poses. Manifold
    Sensing Attention first aggregates each view using the local electrode
    layout and tangent-frame geometry to produce per-view sensing features.
    These features with different poses are fused to produce shape tokens which represents the shape feature, it is then cross-attend to a structured 3D latent, which is converted into the final mesh by a pretrained shape decoder.}
    \Description{The pipeline is arranged from left to right in three connected stages. The left panel shows multiple sensing-manifold poses around an object and the corresponding capacitive fields for views 1 through V. In the center, a magnified electrode neighborhood feeds channel embeddings into Manifold Sensing Attention; the resulting per-view features are combined with their poses by a transformer encoder and cross-attended to a structured 3D latent representation. The right panel shows this latent volume entering a prior-guided shape decoder to produce the final mesh.}
    \label{fig:method_overview}
\end{figure*}
We initialize each channel feature from its capacitive proximity field and a per-channel geometry descriptor that exposes the local electrode information to the network. Let
\(\rho(R)=[t^1,t^2]\in\mathbb R^6\) denote the continuous 6D representation ~\cite{zhou2019continuity} of 3D rotations, and let
\(\chi_j=[x_j,\rho(R_j)]\) be the channel geometry descriptor at site \(j\). The
initial per-channel feature is formulated as:
\begin{equation}
   h_{ij}=\psi_c(C_{ij})+\psi_g(\chi_j)\in\mathbb R^d,
\end{equation}
where the learnable maps \(\psi_c\) and \(\psi_g\) lift the scalar response and
the geometry descriptor to a common dimension \(d\). We collect these features into the per-view feature matrix
\(\mathbf H_i\in\mathbb R^{N\times d}\) for view \(i\). Rather than relying
solely on feature similarity, we then refine \(\mathbf H_i\) with
manifold sensing attention on the local electrode neighborhood, which provides a consistent parameterization of the local chart on the sensing manifold.

For a local channel pair \((j,k)\) on the manifold, we map electrode $k$ to the local
tangent frame of electrode $j$. Specifically, the local geometry descriptor
\(\delta_{jk}=\bigl[R_j^\top(x_k-x_j),\rho(R_j^\top R_k)\bigr]\) encodes local
displacement and relative frame orientation with respect to this tangent space. A learnable map $\psi_m$ then maps the local
descriptor to a local manifold sensing embedding
\begin{equation}
m_{jk}=\!\psi_m(\delta_{jk})\in\mathbb R^d.
\end{equation}

Neighborhood attention then injects \(m_{jk}\) into the key and value of neighbor \(k \in \mathcal N(j)\), where $\mathcal N(j)$ denotes the neighborhood of
electrode $j$ from the adjacency graph $\mathcal G$. Consequently, local geometry and connectivity affect the attention weights and per-channel sensing features:
\begin{align}
a_{ijk}
&=
\operatorname*{softmax}_{k\in\mathcal N(j)}
\!\left(
\frac{(\mathbf W_Q h_{ij})^{\!\top}(\mathbf W_K h_{ik}+m_{jk})}
     {\sqrt d}
\right), \notag\\
h_{ij}
&\leftarrow h_{ij}+\sum_{k\in\mathcal N(j)}a_{ijk}\,
   \bigl(\mathbf W_V h_{ik}+m_{jk}\bigr).
\label{eq:intrinsic_update}
\end{align}
Here
\(\mathbf W_Q,\mathbf W_K,\mathbf W_V\in\mathbb R^{d\times d}\) are learned
query, key and value weights\\~\cite{vaswani2017attention}.

Finally, for each channel, we compute a sensor-shape embedding
\begin{equation}
    s_j=\psi_s(\chi_j) \in\mathbb R^d,
\end{equation}
learned-query pooling $\Phi$ over channels yields the per-view feature vector $u_i$, where 
\begin{equation}
   u_i = \Phi\left(\{[h_{ij}, s_j]\}_{j=1}^{N}\right) \in\mathbb R^d.
\end{equation}

\subsection{Reconstruction with Shape Tokens}

\paragraph{Multi-view Fusion via Shape Tokens.}

To extract the underlying object's 3D shape from viewpoint-dependent sensing features, we aggregate these per-view sensing features into shape tokens.

We therefore concatenate $u_i$ with a Fourier encoding of the per-view pose descriptor
$\xi_i = [b_i,\rho(\bar R_i)]$,
\begin{equation}
    \tilde u_i \;=\; \psi_T\!\bigl([\,u_i,\, \gamma(\xi_i)\,]\bigr) \in \mathbb R^{d},
\end{equation}
where $\gamma$ is sinusoidal Fourier
encoding~\cite{mildenhall2020nerf}. We introduce
8 learnable shape tokens, empirically chosen to balance expressiveness against overfitting, as a shape-level representation of the
multi-view capacitive proximity fields. These tokens are prepended to the per-view sensing feature, together with 4 register tokens~\cite{darcet2024registers,oquab2024dinov2}
that provide non-semantic capacity for the transformer and reduce
spurious accumulation in the shape token. 

Through self-attention, the multi-view sensing features exchange cross-view context, whereby the shape tokens can selectively aggregate global evidence from the entire multi-view sequence. The resulting shape tokens \(\mathbf S \in \mathbb R^{8\times d}\) therefore provide a compact representation of object's 3D shape.

\paragraph{Shape Decoding with Prior.}

Each capacitive proximity field contains only \(N\) electrode responses,
typically \(100\)--\(300\), from a single scan view. While multiple fields
provide complementary evidence, reconstruction remains challenging due to this
sparsity. We therefore leverage a pre-trained TRELLIS-2 shape decoder \cite{xiang2025trellis2}, whose learned shape prior facilitates the reconstruction of complete object geometry.

First, we map the shape tokens to the 3D latent structure required by the shape prior. Specifically, the target shape decoder operates on a
subset of an \(8^3\) resolution latent grid, with a 32-dimensional feature per occupied
cell. Let \(\mathcal I=\{0,\ldots,7\}^3\) be the latent-cell index set. 

For each latent cell \(\ell \in \mathcal I\), a learned query \(q_\ell\) cross-attends to the
shape tokens, \(r_\ell=\operatorname{CrossAttn}(q_\ell,\mathbf S)\). A shared
per-cell map $\psi_\mathrm{c}$ predicts
\((\hat s_\ell,\hat f_\ell)=\psi_{\mathrm{c}}(r_\ell)\), where
\(\hat s_\ell\in[0,1]\) is the occupancy probability and
\(\hat f_\ell\in\mathbb R^{32}\) is the per-cell latent feature. Across the 3D latent grid, the occupancy probabilities \(\{\hat s_\ell\}\) act as a gate: cells whose probability exceeds 0.5 are retained to form the active set \(\hat{\mathcal A}=\{\ell\in\mathcal I:\hat s_\ell>0.5\}\), while the rest are pruned away. The features attached to the active cells are collated into the sparse latent \(\hat z=\{(\ell,\hat f_\ell):\ell\in\hat{\mathcal A}\}\), which directly matches the input format required by the target shape decoder.

The shape decoder \(D\) then synthesizes the final mesh
\(\hat M=D(\hat z)\) in the canonical object frame.

\subsection{Training}
\label{sec:training}

Each training sample pairs a set of \(V\) posed capacitive proximity fields \(\{(C_i,\bar R_i,b_i)\}_{i=1}^{V}\), observed on the sensor geometry
\((\{x_j,R_j\}_{j=1}^{N},\mathcal G)\), with a ground-truth object
mesh \(M\) in the canonical object frame. The framework discussed above predicts a sparse latent \(\hat z\) from these posed fields,
while the frozen TRELLIS-2 encoder encodes the mesh \(M\) into
the supervision target
\(z^\star=\{(\ell,f_\ell^\star):\ell\in\mathcal A^\star\}\), where
\(\mathcal A^\star\subseteq\mathcal I\), and the occupancy probability target is
\(s_\ell^\star=\mathbbm{1}[\ell\in\mathcal A^\star]\).

Training supervises occupancy probabilities, cell features, and decoder-side
consistency with different weights $\lambda$:
\begin{align}
\mathcal L
&= \lambda_o \mathcal L_o
 + \lambda_f \mathcal L_f
 + \lambda_D \mathcal L_D, \label{eq:training_loss}\\
\mathcal L_o
&= \mathcal L_{\mathrm{SoftIoU}}(\hat s,s^\star), \notag\\
\mathcal L_f
&= \frac{1}{|\mathcal A^\star|}
   \sum_{\ell\in\mathcal A^\star}
   \|\hat f_\ell-f_\ell^\star\|_2^2, \notag\\
\mathcal L_D
&= \Pi_D\!\left(
   D(\hat z), M
   \right). \notag
\end{align}
\(\Pi_D\) is the Chamfer-L1 loss between sampled
surface points on \(D(\hat z)\) and \(M\). Note that \(\mathcal L_D\) acts as a training-time regularizer; the TRELLIS-2 decoder remains frozen throughout training.

\section{Surrogate Forward Model for Capacitive Proximity Fields}
\label{sec:neural_forward_sim}

Physical data acquisition is time consuming at scale. To mitigate this, we formulate data generation as a generative problem. Within the simulated environment, each channel's response is approximated via hemispherical raycasting in its local tangent frame. Specifically, for a channel site $x_j$,  $K$ random directions $\{r_{jk}\}_{k=1}^K$ are sampled from the upper-hemisphere $\mathbb S^2_+$. Along each direction $r_{jk}$, the first-hit distance $d_{jk}$ is computed to the target surface, which is clipped by a maximum near-field range $d_{\max}=30mm$.

The \emph{sensing geometric prior} $g_j$ is defined as a cosine-weighted fraction of this hemisphere:
\begin{equation}
        g_j = \frac{ \sum_{k=1}^{K} \omega_{jk} \left(1-\frac{d_{jk}}{d_{\max}}\right) }{ \sum_{k=1}^{K} \omega_{jk} }.
\end{equation}

Here, $\omega_{jk} = r_{jk}^{\top}n_j$ measures the alignment with the local normal $n_j$ to account for the directional dependency over the sensing manifold.

To map the geometric prior to physical response, we learn a surrogate model $p_\theta(\mathbf{C}\mid g)$ via a conditional Diffusion Transformer (DiT)~\cite{peebles2023scalable}. Treating the sensing geometric prior $g$ as a guiding condition, the network is trained via standard noise-prediction:

\begin{equation}
    \mathcal{L}_{\text{sim}} = \mathbb{E}_{\mathbf{C}, \epsilon, t} \left[ \| \epsilon - \epsilon_\theta(\mathbf{C}_t, g, t) \|_2^2 \right],
\end{equation}
where $t$ is the uniformly sampled diffusion timestep, $\mathbf{C}_t$ is the noisy capacitive field corrupted by the algorithmic Gaussian noise $\epsilon \sim \mathcal{N}(0, \mathbf{I})$, and $\epsilon_\theta$ is our surrogate model. 

Notably, we equip the denoising network $\epsilon_\theta$ with the same \emph{Manifold Sensing Attention} (MSA) module introduced in our reconstruction pipeline. By explicitly injecting local geometry embeddings into the self-attention layers, MSA grounds the generative process directly in the underlying local geometry of the sensing manifold.

In practice, we first pre-train the surrogate model on a large synthetic dataset generated by FEM simulations. This provides the network with a strong physical prior. Subsequently, we fine-tune the model exclusively on the small-scale, physical dataset collection. During this fine-tuning phase, the generative capacity of the diffusion process adapts the FEM physical prior to the true distribution of the real world hardware. 

\section{Experiments}
\label{sec:experiments}

As illustrated in Fig.~\ref{fig:hemisphere}, our physical experimental setup utilizes a hemispherical sensing manifold equipped with 100 electrode channels, with a TM-500 robotic arm controlling the spatial pose.

\subsection{Experimental Setup}
\label{sec:evaluation_setup}

\paragraph{Datasets.}
Our training data consists of 500 FEM-simulated meshes and 25 physical objects, each sampled at 512 poses. We trained a surrogate forward model mentioned in Sec.\ref{sec:neural_forward_sim} on the simulation data and fine-tuned it using the measurements from physical objects to align with real-world proximity response distribution. The resulting model is used to generate realistic proximity data for a larger set of 1,624 meshes (512 views each). Apart from the physical objects, all mesh data are sampled from standard datasets, including Google Scanned Objects \cite{downs2022google}, Manifold40 \cite{hu2022subdivnet}, and Thingi10K \cite{zhou2016thingi10k}.

Additional implementation details are provided in the supplemental material.

\paragraph{Metrics.}
Inspired by TouchSDF \cite{comi2023touchsdf}, our evaluation utilizes Chamfer Distance (CD), Earth Mover’s Distance (EMD), and Average Surface Error which quantifies the average distance between the predicted surface $\mathcal{S}_p$ and the ground-truth $\mathcal{S}_{gt}$. Although minimized during training, CD is often biased toward local point density. Thus, we incorporate EMD and Average Surface Error to comprehensively evaluate the global structural fidelity.

\subsection{Shape Reconstruction}
\label{sec:multiview_reconstruction}

\begin{figure*}[t]
    \centering
    \includegraphics[width=\textwidth]{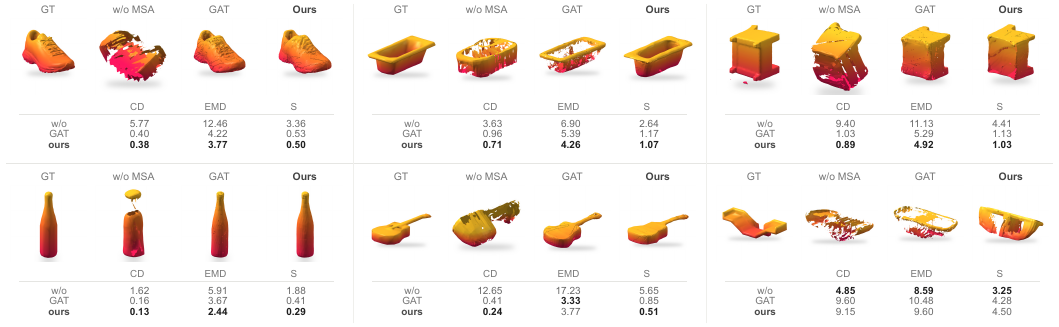}
    \caption{Ablation comparing vanilla self-attention (``w/o MSA''), GAT, and MSA (Ours) under identical inputs. GAT and MSA use the same channel adjacency, whereas ``w/o MSA'' attends across all electrode channels. MSA achieves the best reconstruction fidelity in most cases. The example in the bottom-right corner shows a failure case caused by insufficient capacitive evidence. Metrics are scaled for compact display: $CD \times 10^3$, $EMD \times 10^2$, and S (Av. Surface err.) $\times 10^2$; lower is better.}
    \Description{Six examples are arranged in a two-by-three grid. Within each example, four aligned columns show the ground-truth shape followed by reconstructions without MSA, with GAT, and with MSA; a table directly below reports the three error measures for the same methods. The vertical alignment supports method-by-method comparison against the ground truth, showing that the MSA reconstruction usually preserves the reference geometry more completely, while all methods degrade on the final example at the lower right.}
    \label{fig:ablation}
\end{figure*}

Shape reconstruction performance is evaluated on 200 unseen test shapes by using generated proximity data. For each shape, we iteratively sample random poses to construct 512 valid views (i.e., those with non-zero sensor responses). Our standard evaluation utilizes shape reconstruction via a random subset of 64 views per object. To rigorously assess our framework, we conduct two further analyses. First, to validate the importance of Manifold Sensing Attention (MSA), we compare it against vanilla self-attention and GAT~\cite{velickovic2018graph}, which also encodes channel adjacency. Second, we investigate the impact of view density by varying the input view count $V \in \{8, 16, 32, 64, 96, 128\}$, where each V is random drawn from the 512 views.

\begin{table}[!ht]
    \centering
    \small
    \setlength{\tabcolsep}{5pt}
    \caption{Quantitative shape reconstruction results: An ablation study validating the Manifold Sensing Attention (MSA) module against vanilla self-attention and GAT baselines and an investigation into the impact of varying input view counts.}
    \begin{tabular*}{\columnwidth}{@{\extracolsep{\fill}}llccc@{}}
        \toprule
        Study & Setting & CD $\downarrow$ & EMD $\downarrow$ & Av. Surf. Err. $\downarrow$ \\
        \midrule
        MSA ablation & w/o MSA & 0.0820 & 0.0985 & 0.0217 \\
                     & GAT & 0.0185 & 0.0693 & 0.0158 \\
                     & w/ MSA  & 0.0078 & 0.0495 & 0.0071 \\
        \midrule
        Views & 8  & 0.0286 & 0.1932 & 0.0458 \\
        & 16  & 0.0169 & 0.1536 & 0.0301 \\
                   & 32  & 0.0098 & 0.0506 & 0.0103 \\
                   & 64  & 0.0078 & 0.0495 & 0.0071 \\
                   & 96  & 0.0077 & 0.0466 & 0.0073 \\
                   & 128 & 0.0079 & 0.0493 & 0.0074 \\
        \bottomrule
    \end{tabular*}
    \label{tab:reconstruction_results}
\end{table}

Table~\ref{tab:reconstruction_results} shows that MSA outperforms both baselines. Reconstruction improves through $V=64$ and then plateaus, so we use $V=64$ by default to balance reconstruction fidelity and computational efficiency.

\begin{figure}[t]
    \centering
    \includegraphics[width=0.49\textwidth]{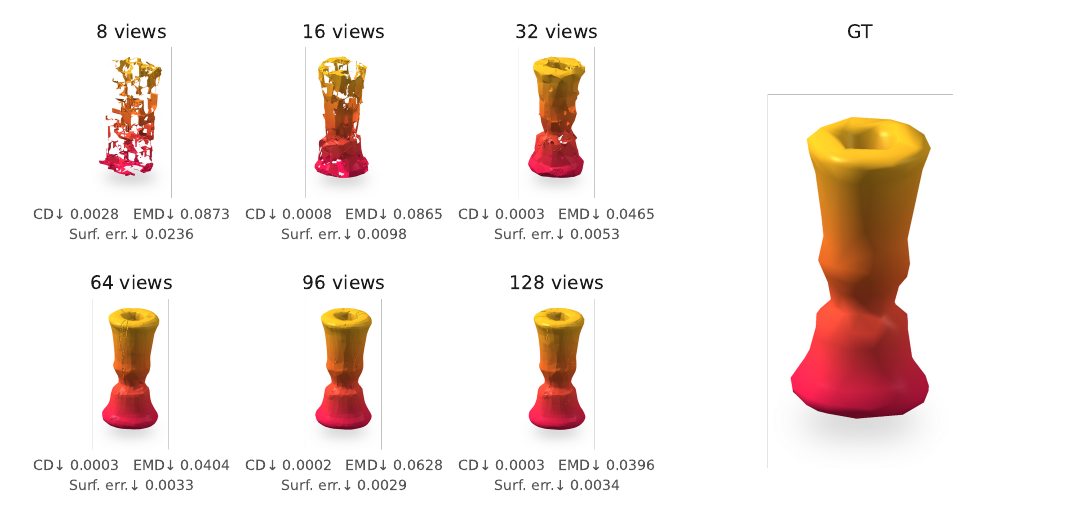}
    \caption{Shape Reconstruction from different numbers of input views. We investigate the impact of view density on reconstruction performance by varying the input view count $\{8, 16, 32, 64, 96, 128\}$.}
    \Description{The left side contains a two-by-three grid of reconstructions using 8, 16, 32, 64, 96, and 128 views, with three error values below each result; a larger ground-truth model appears on the right for reference. Reading the reconstruction panels in increasing view count shows the shape changing from fragmented at 8 views to largely complete by 32 views and visually stable from 64 through 128 views.}
    \label{fig:view_counts}
\end{figure}

The visualizations of the reconstruction results for the MSA ablation and varying view counts are presented in Fig.~\ref{fig:ablation} and Fig.~\ref{fig:view_counts}, respectively. Additional reconstruction results from simulated proximity signals are shown in Fig.~\ref{fig:recons}.

\subsection{Surrogate Forward Model Validation}
\label{sec:surrogate_validation}

We validate the surrogate forward model on held-out physical object-pose pairs that are not used during real-data fine-tuning. For each sample, we compare the generated capacitive proximity field against the measured hardware response over all sensing channels. Table~\ref{tab:surrogate_validation} reports MAE and RMSE to quantify absolute response errors, and reports Pearson and Spearman correlation to test whether the surrogate preserves the channel-wise response pattern induced by each object pose.

\begin{table}[!ht]
    \centering
    \small
    \setlength{\tabcolsep}{5pt}
    \caption{Surrogate forward-model validation on held-out physical measurements. Metrics are computed channel-wise between generated and measured capacitive proximity fields; PCC and SRC denote Pearson and Spearman correlation, respectively.}
    \begin{tabular*}{\columnwidth}{@{\extracolsep{\fill}}lcccc@{}}
        \toprule
        Model & MAE $\downarrow$ & RMSE $\downarrow$ & PCC $\uparrow$ & SRC $\uparrow$ \\
        \midrule
        FEM only & $9.18\times 10^{-7}$ & $8.23\times10^{-4}$ & 0.9852 & 0.9659 \\
        FEM + real & $1.28\times 10^{-7}$ & $1.81\times10^{-4}$ & 0.9990 & 0.9973 \\
        \bottomrule
    \end{tabular*}
    \label{tab:surrogate_validation}
\end{table}

\section{Application: Dexterous Hand with Palm-Mounted Sensing Manifold}
\label{sec:application}

We evaluate whether the same reconstruction pipeline can be used when the
sensing manifold is integrated into a manipulation platform. Specifically, we
attach a woven-based capacitive sensing manifold to the palm side
of a RH56F1 dexterous hand from Inspire Robots and use the hand motion to acquire multi-view proximity signals of an object and then reconstruct object's shape to facilitate grasp planning.

\paragraph{Proximity data collection on palm sensor.}
The palm sensor is a calibrated sensing manifold in the hand frame. Specifically, each capacitive channel has a known channel site, local electrode frame, and adjacency relation after the fabrication. During the scanning process, we randomly sample hand poses around the object and use the proximity response itself as a pre-contact safety signal. A candidate pose is accepted as a valid view when the maximum channel response in the sensing manifold lies within a prespecified proximity band: if the response is below the lower
threshold, the hand moves closer to the object; if it exceeds the upper
threshold, the hand retreats before contact and resamples. 

\paragraph{Reconstruction.}
We retrain the Proximity3D pipeline using
the same object corpus as the hemispherical sensing manifold, consisting of
1,624 meshes with generated multi-view capacitive proximity fields. The
training step is identical to that for the hemispherical sensing manifold except that the sensing manifold is replaced by the hand-palm geometry. In this way, the network receives the
same type of input representation---capacitive proximity fields, known sensor
poses, channel sites, local electrode frames, and channel adjacency---while
adapting the local aggregation to the palm-shaped manifold.

We set \(V=64\) as the default number of input views, following the view-count
calibration. At inference time, the \(V\) accepted
pre-contact scans from the random pose sampling procedure are fed
to reconstruct the object mesh in the object frame.
Fig.~\ref{fig:palm_application_reconstruction} shows three representative
reconstruction results from this setting. These real world cases illustrate that the palm-mounted sensing manifold can provide sufficient near-field observations for object shape recovery to facilitate grasp planning.

\paragraph{Grasp planning evaluation.}
To test how reconstructed geometry improves downstream grasp planning, we use GraspGen \cite{murali2026graspgen} to plan grasps on our reconstructions and on the ground-truth convex hull, oriented bounding box (OBB), and axis-aligned bounding box (AABB). Each grasp is evaluated against the ground-truth mesh using a standard force-closure check \cite{ferrari1992planning}. We evaluate 100 objects for which planning on the ground-truth mesh succeeds, so failures reflect the geometric input alone. Our reconstructions achieve a 90\% success rate, compared with 78\%, 68\%, and 65\% for the convex hull, OBB, and AABB, respectively.

\section{Discussion}
\label{sec:discussion}

\begin{figure*}[t]
    \centering
    \includegraphics[width=0.85\textwidth]{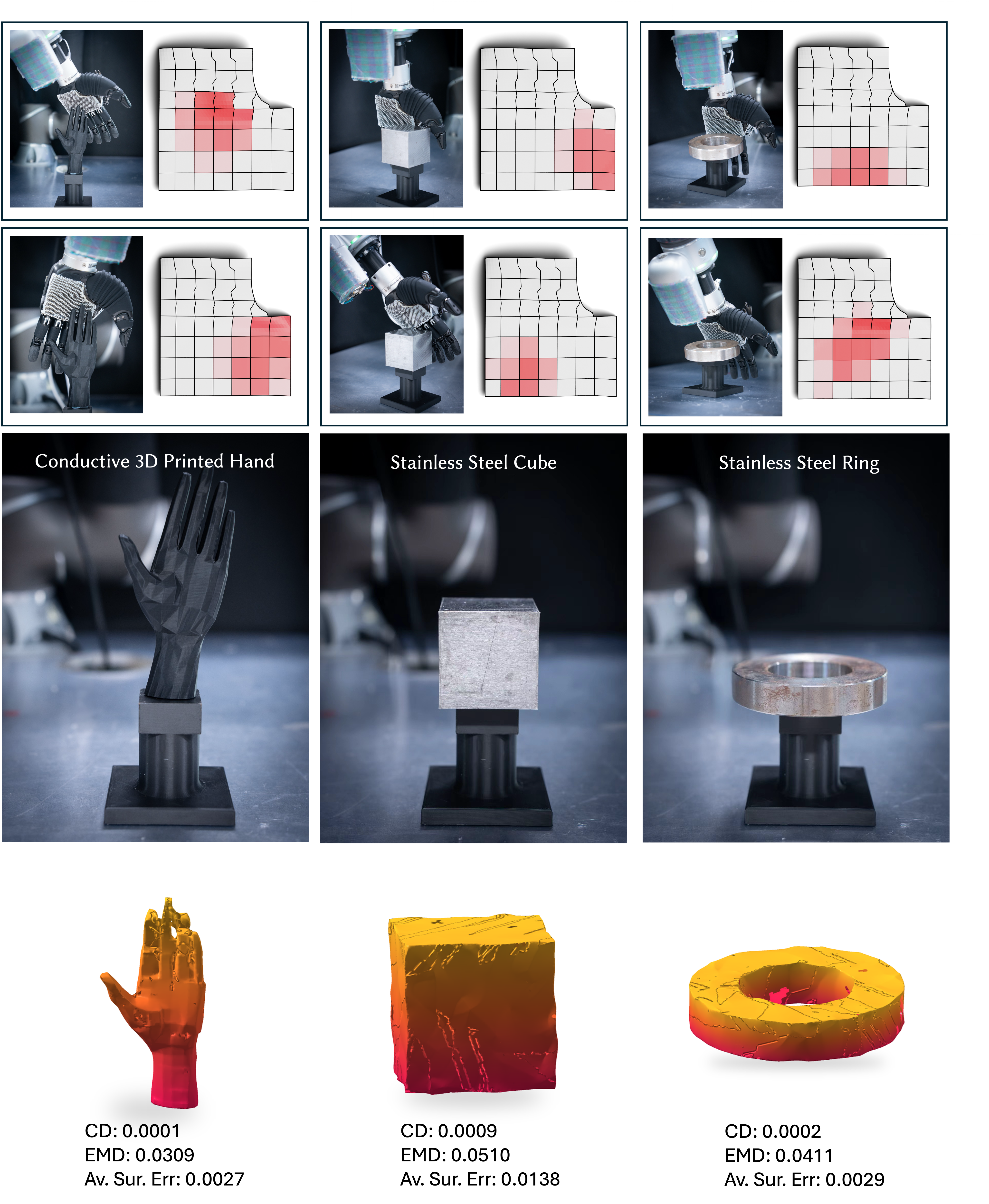}
    \caption{Illustration of the woven-based sensing manifold mounted on a dexterous hand's palm part. Each column displays the physical scanning process alongside the final reconstruction result. From left to right: a conductive 3D-printed hand model, and a cube and ring both fabricated from stainless steel.}
    \Description{Three columns present the hand, cube, and ring experiments in parallel. At the top of each column, two photographs of the palm-mounted sensor approaching the object are paired with red proximity-field maps for the corresponding poses. The middle row shows the physical target, and the bottom row shows its reconstructed surface with error metrics, creating a direct top-to-bottom link from multi-view measurements to target and reconstruction.}
    \label{fig:palm_application_reconstruction}
\end{figure*}

\begin{figure*}[t]
    \centering
    \includegraphics[width=0.85\textwidth]{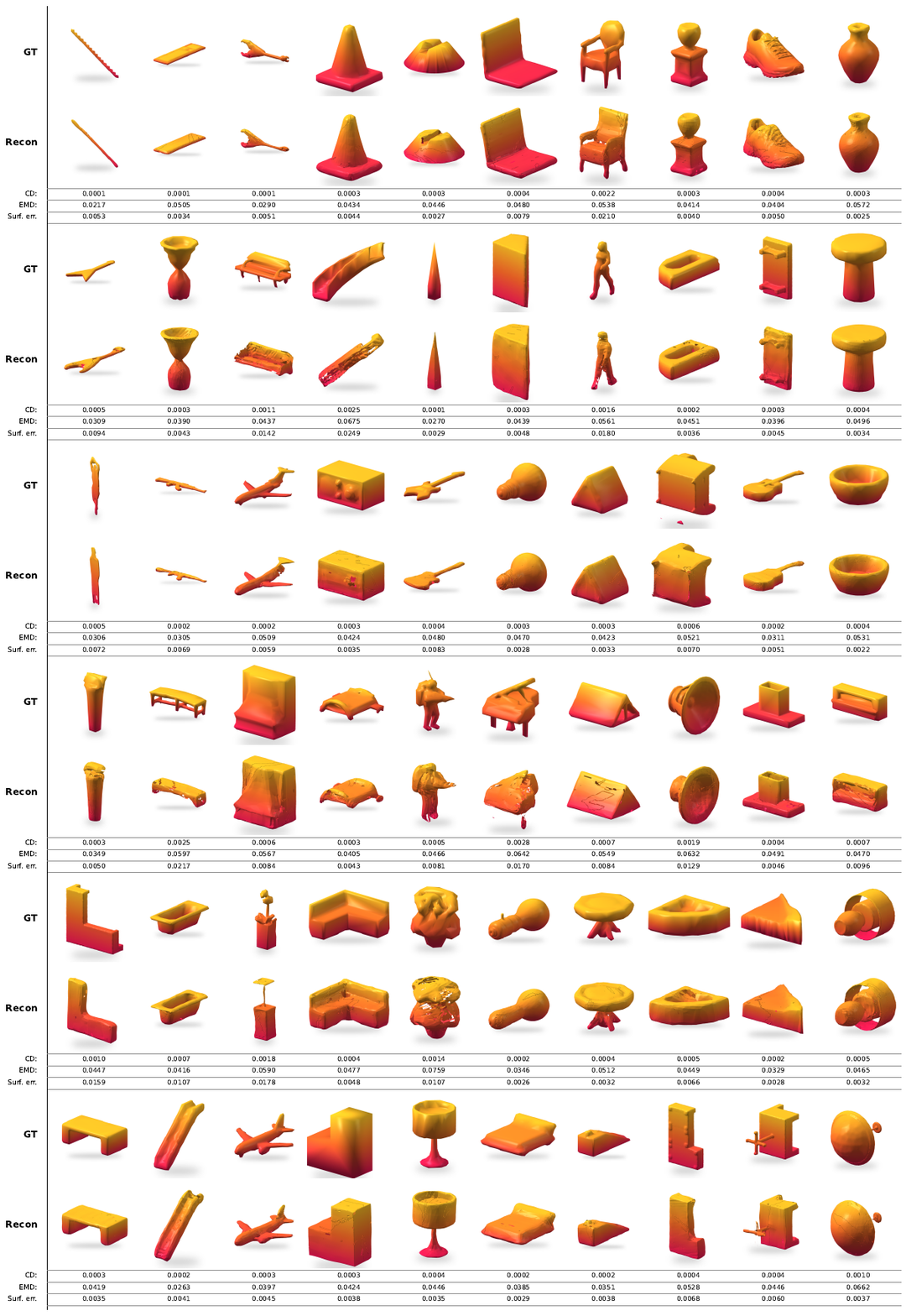}
    \caption{Reconstruction results using simulated capacitive proximity fields on the hemispherical sensing manifold.}
    \Description{A gallery of sixty test shapes is divided into six horizontal blocks. Each block pairs a row of ground-truth shapes with a row of reconstructions directly beneath them, so every reconstructed object is vertically aligned with its reference; three error values appear below each pair. The repeated layout enables comparison across thin, branched, concave, rounded, and box-like geometries and reveals both close matches and localized missing or distorted parts.}
    \label{fig:recons}
\end{figure*}

Proximity3D targets 3D shape recovery from multi-view capacitive proximity fields on sensing manifolds, but its applicability is currently limited to conductive objects because the sensing mechanism relies on electric-field coupling. Within this regime, material primarily changes the overall coupling strength while exerting a weaker influence on the normalized spatial patterns used by the network. To quantify this effect, we use a model trained only on stainless-steel-generated data to reconstruct the same 50 objects made of gold, silver, copper, and aluminum. Relative to stainless steel, CD changes by at most 8.1\% and EMD by at most 6.6\%. We further examine sensitivity to internal structure by replacing a solid stainless-steel sphere of 5\,cm radius with a 1\,cm-thick hollow shell, which changes CD by 1.4\% and EMD by 3.3\%. The conductive-fiber composite hand in Fig.~\ref{fig:palm_application_reconstruction}, whose conductivity is substantially lower than that of the tested metals, is also reconstructed without material-specific retraining. Together, these results suggest robustness across the tested conductive objects, although broader material characterization is needed to determine the method's practical conductivity boundary.

The pretrained decoder provides a useful shape prior that converts sparse multi-view evidence into complete meshes. However, it may also bias reconstruction when the sensed evidence is weak. We therefore assess whether the output remains driven by the sensed evidence rather than by the decoder alone. Two results support this: quality improves with additional views, while replacing MSA with vanilla self-attention degrades performance despite an unchanged decoder (Table~\ref{tab:reconstruction_results}). On 30 held-out OOD shapes with deep concavities or high genus, CD increases by 63.2\% over the in-distribution set, yet the CD to the nearest training shape remains 3.9$\times$ that to the ground truth, indicating graceful degradation rather than prior collapse. This limitation remains visible when capacitive evidence is insufficient (see the bottom-right example in Fig.~\ref{fig:ablation}). Future work should optimize scanning to ensure sufficient information capture.

Weak sensing is not the only potential source of error; approximation by the surrogate model may also contribute to reconstruction failures. Although the surrogate agrees closely with held-out physical measurements, the current evaluation does not quantify how much of the observed reconstruction error is attributable to surrogate approximation. Future work will estimate this contribution using paired physical and surrogate measurements under matched sensing configurations.

In summary, Proximity3D presents a pioneering framework for recovering 3D object geometry from multi-view capacitive proximity fields captured across a curved sensing manifold. Validated through physical experiments, this approach successfully demonstrates pre-contact geometric awareness and marking a significant step forward for embodied near-field sensing, with the potential to enable a broader range of robotic applications.

\begin{acks}
This study was supported by the Centre for Perceptual and Interactive Intelligence,
a CUHK-led InnoCentre under the InnoHK initiative of the Innovation and Technology
Commission of the Hong Kong Special Administrative Region Government.
\end{acks}

\bibliographystyle{ACM-Reference-Format}
\bibliography{references}


\clearpage
\onecolumn
\suppressfloats[t]

\begin{center}
    {\LARGE\bfseries Supplementary Material}
\end{center}

\setcounter{section}{0}
\renewcommand{\thesection}{\Alph{section}}
\renewcommand{\theHsection}{supplement.\arabic{section}}
\setcounter{figure}{0}
\renewcommand{\thefigure}{S\arabic{figure}}
\renewcommand{\theHfigure}{supplement.\arabic{figure}}
\setcounter{equation}{0}
\renewcommand{\theequation}{S\arabic{equation}}
\renewcommand{\theHequation}{supplement.\arabic{equation}}

\section{Woven-based Capacitive Sensing Manifold}

\subsection{Sensor Architecture and Readout}
\label{sec:sensor_architecture}

The fabric interlaces two conductive yarn systems. Bare stainless-steel fibers
run along the warp direction and form the sensing electrodes; silicone-insulated
copper wires run along the weft direction and form the driven electrodes. The
silicone coating prevents electrical contact at warp--weft crossings while
allowing capacitive coupling. Several adjacent conductive threads are grouped
into one channel. Thus, an individual thread crossing is only a structural cross
point: a \emph{logical sensing site} is addressed by one grouped warp channel and
one grouped weft channel. The group size trades spatial resolution against
routing and readout complexity.

\begin{figure*}[t]
    \centering
    \includegraphics[width=0.97\textwidth]{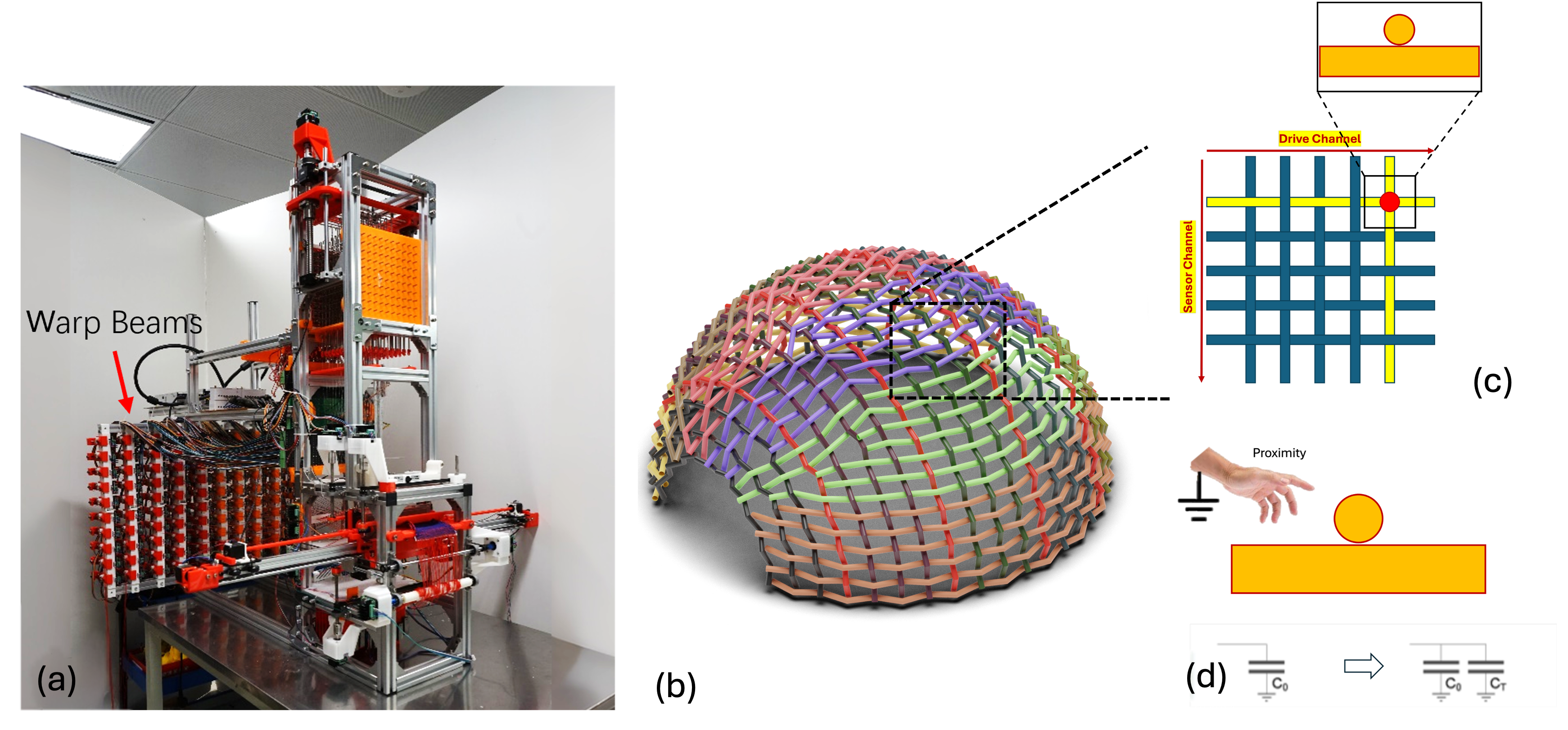}
    \caption{
    (a) The computer-controlled freeform
    weaving system. (b) The woven
    hemispherical sensing manifold. (c) Illustration of a drive- and sensor-channel pair. (d) Proximity sensing model.
    }
    \Description{A four-panel figure showing the freeform weaving machine, woven hemispherical sensor, multiplexed drive and sensor channels, and the equivalent capacitance change caused by an approaching target.}
    \label{fig:hardware_sensor}
\end{figure*}

The readout electronics time-multiplex the channel pairs shown in
Fig.~\ref{fig:hardware_sensor}(c). For each addressed site, the selected weft
channel is excited and the charge transferred to the selected warp channel is
integrated to estimate capacitance. Repeating the scan over all sites produces
one spatially indexed capacitance vector while preserving the correspondence
between every scalar measurement and its physical location on the curved
surface.

\subsection{Proximity Sensing Principle}
\label{sec:proximity_sensing_principle}

At site $j$, the driven and sensing electrodes form a baseline mutual
capacitance $C_{0,j}$ that includes the intended warp--weft coupling and fixed
parasitic contributions from the textile and wiring. As illustrated in
Fig.~\ref{fig:hardware_sensor}(d), a nearby conductive object introduces an
additional capacitive path to the electrodes and their electrical environment.
This path redistributes the fringing electric field and changes the charge
received by the sensing electrode, following the same electric-field
perturbation used in capacitive pretouch sensing~\cite{mayton2010electric}. Let
$\tilde C_{ij}$ be the raw capacitance at site $j$ in view $i$. The signal
response used in the main paper is
\begin{equation}
    C_{ij}=\operatorname{Norm}\!\left(
    \left\lvert\tilde C_{ij}-C_{0,j}\right\rvert\right),
    \label{eq:baseline_response}
\end{equation}

For intuition, the magnitude of the object--electrode coupling increases with
effective coupled area and decreases with separation,
\begin{equation}
    \left\lvert \tilde C_{ij}-C_{0,j} \right\rvert \;\propto\;
    \varepsilon_0\varepsilon_r
    \frac{A_{\mathrm{eff},ij}}{d_{ij}}.
    \label{eq:proximity_intuition}
\end{equation}
Here, $\varepsilon_0$ is the vacuum permittivity, $\varepsilon_r$ is the
relative permittivity of the intervening medium, $A_{\mathrm{eff},ij}$ is the
effective coupled area, and $d_{ij}$ is the object--electrode separation.
This relation is only a local approximation: curved electrodes, fringing fields,
grounding, routing parasitics, and neighboring channels make
the actual response nonlinear and spatially distributed. A single channel is
therefore not treated as a direct depth measurement. After baseline correction
and calibration, the site responses form
$C_i=[C_{i1},\ldots,C_{iN}]$ on the sensing manifold.
\subsection{Sensor Fabrication}
\label{sec:sensing_manifold_fabrication}

We derive the textile layout from the geodesic stitch-map formulation for 4D
garment knitting~\cite{liu2021knitting4d} and fabricate it using the
computer-controlled 3D freeform weaving system shown in
Fig.~\ref{fig:hardware_sensor}(a)~\cite{chen2024freeformweaving}.
The target surface is compiled into machine instructions: the Jacquard mechanism
selects the warp threads for each weft pass, the independently controlled warp
beams release the local thread lengths required to form curvature, and the
shuttle inserts the weft, producing the manifold shown in
Fig.~\ref{fig:hardware_sensor}(b).

\section{Implementation Details}
\label{sec:implementation_details}

Unless stated otherwise, reconstruction uses \(V{=}64\) posed proximity fields
per object. The reconstruction network contains two MSA layers for local
manifold aggregation, a six-layer Transformer for multi-view fusion, and
a four-layer cross-attention cell-query decoder for 3D latent projection.
The feature width is \(d{=}384\), and \(\psi_c, \psi_g, \psi_s, \psi_m, \psi_T\), and
\(\psi_{cm}\) are implemented as two-layer MLPs. The reconstruction loss weights are
\(\lambda_o=1.0\), \(\lambda_f=1.0\), and \(\lambda_D=0.1\) for
the training loss defined in the main paper. The TRELLIS-2 shape decoder
is kept frozen during training and inference.

For the surrogate forward model, \(K{=}512\) random directions are sampled from
the local upper hemisphere \(\mathbb S^2_+\) at each channel to construct the
geometric conditioning signal.

Gradient computations were performed using the PyTorch framework. All trainable modules use AdamW with a base learning rate of \(1{\times}10^{-4}\), \(0.05\) weight decay, gradient clipping at \(1.0\), a \(2,000\)-step linear warmup, and cosine decay. Training is
conducted on a single node with \(8{\times}\) NVIDIA RTX 4090 GPUs. The source code will be released upon the acceptance of this paper.

\section{Architecture Generalization}
\label{sec:architecture_generalization}

\begin{figure*}[t]
    \centering
    \includegraphics[width=0.98\textwidth]{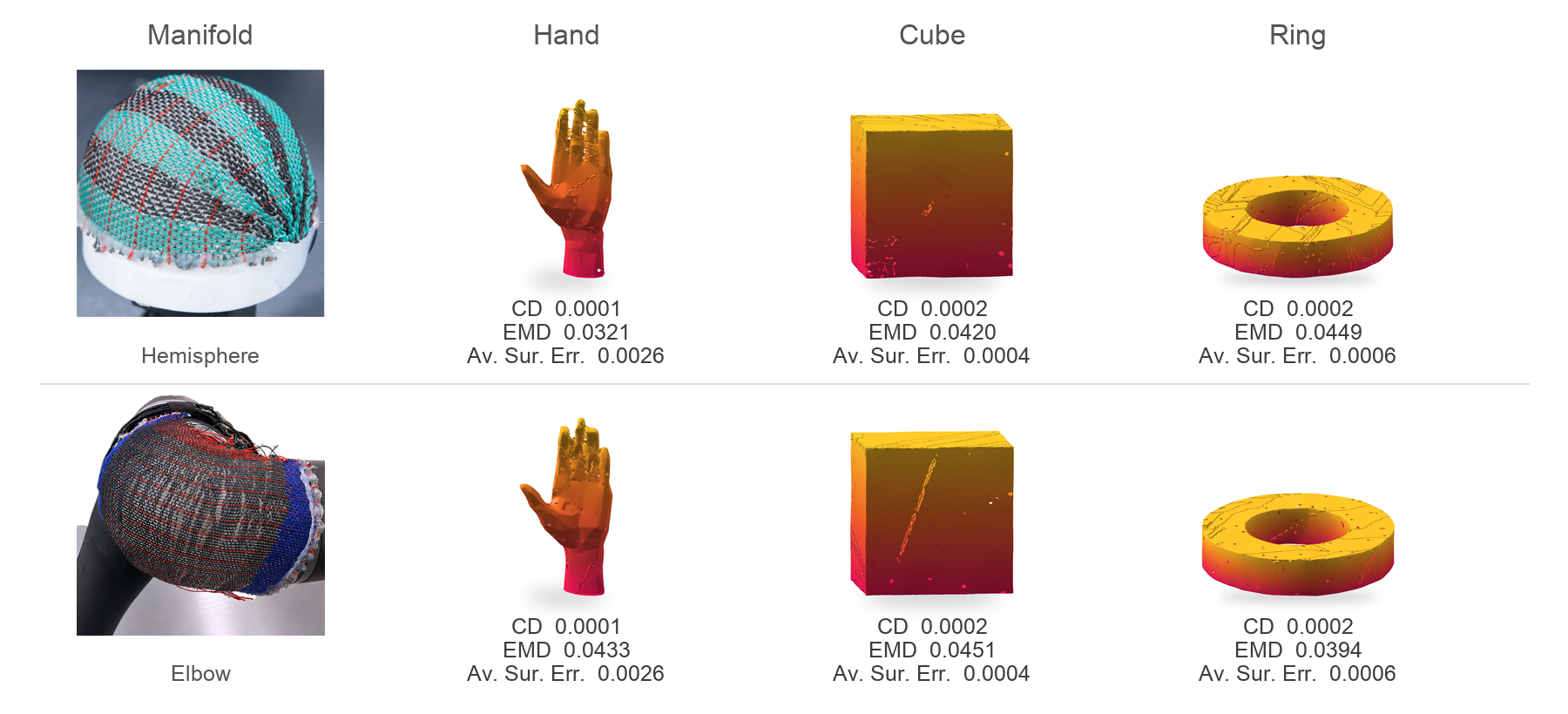}
    \caption{Reconstruction of the same hand, cube, and ring using sensors
    conforming to a hemisphere and a robotic-arm elbow.}
    \Description{A comparison of hand, cube, and ring reconstructions obtained
    with hemispherical and elbow-mounted sensing manifolds, with photographs of
    both physical sensing manifolds in the leftmost column.}
    \label{fig:manifold_generalization}
\end{figure*}

To evaluate architecture generalization, we reconstruct the same hand, cube, and
ring using sensing manifolds conforming to three distinct geometries: a
hemisphere, the elbow of a robotic arm, and the palm of a dexterous hand. Both the surrogate and
reconstruction models are retrained for each manifold using its own geometry and sensor layout, while retaining the same
architectures and training settings. The ground-truth shapes and reconstructions
obtained with the palm-mounted sensor are shown in Fig.~6 of the main paper;
Fig.~\ref{fig:manifold_generalization} presents the corresponding results on
the other two manifolds. Proximity3D maintains similar reconstruction quality
across the three sensing geometries.

\end{document}